\documentclass[journal]{IEEEtran}
\usepackage{multirow,booktabs,color,soul,threeparttable}
\definecolor{hl}{rgb}{0.75,0.75,0.75}
\sethlcolor{hl}
\usepackage{graphicx}
\usepackage{subfigure}
\usepackage{amsmath}
\usepackage{algpseudocode}
\usepackage{stfloats}
\usepackage{multirow}
\usepackage{verbatim}
\usepackage{array}
\usepackage{marvosym}
\usepackage{soul}
\usepackage{amssymb}
\usepackage[linesnumbered,ruled,vlined]{algorithm2e}
\soulregister\cite7 
\soulregister\citep7 
\soulregister\citet7 
\soulregister\pageref7 

\usepackage[ruled,vlined,linesnumbered]{algorithm2e} 
\usepackage{colortbl}
\usepackage[table,xcdraw]{xcolor}

\begin{document}
	\title{Decoupling Constraints from Two Directions for Evolutionary Constrained Multi-objective Optimization}

	\markboth{IEEE Transactions on Systems, Man and Cybernetics: Systems, \today}
	{Shell \MakeLowercase{\textit{et al.}}: Bare Demo of IEEEtran.cls for IEEE Journals}
	
	\author{
		 Ruiqing Sun, Dawei Feng,  Xing Zhou, Lianghao Li, Sheng Qi, Bo Ding, Yijie Wang,  Rui Wang \IEEEmembership{Senior Member,~IEEE}, Huaimin Wang
   
  	
		\thanks{Ruiqing Sun, Dawei Feng, Bo Ding, Yijie Wang, and Huaimin Wang are with the College of Computer Science and Technology, National University of Defense Technology, Changsha 410000, P.R. China (e-mail: sunny0331@foxmail.com, wangyijie@nudt.edu.cn).
  
        Xing Zhou is with the College of Intelligence Science and Technology, National University of Defense Technology, Changsha, 410000, P.R. China.

        Lianghao Li is with the State Key Laboratory of Complex \& Critical Software Environment, College of Information and Communication, National University of Defense Technology, Wuhan 430019, P.R. China.
  
        Sheng Qi and Rui Wang are with the College of System Engineering, National University of Defense Technology, Changsha 410000, P.R. China.
  }
		}
\maketitle

\begin{abstract}
Real-world constrained multi-objective optimization problems (CMOPs) commonly involve multiple constraints, and understanding and exploiting their coupling relationships is crucial for efficient optimization. Recent constraint-decoupling methods handle individual constraints separately, but they generally search only in the evolutionary direction to approximate single-constraint Pareto fronts (SCPFs). In this study, we show that part or all of the constrained Pareto front (CPF) may be unrelated to any SCPF and instead be shaped by the boundaries of infeasible regions. We refer to such a portion as the independent CPF (ICPF) and introduce the reverse CPF (RCPF) to characterize its associated informative infeasible boundaries. Based on these observations, we propose a bidirectional constraint-decoupling coevolutionary algorithm named DCF2D. DCF2D dynamically identifies the constraints obstructing the main population and activates constraint-specific auxiliary populations. These populations adaptively search in the evolutionary direction for the corresponding SCPFs or in the reverse evolutionary direction for the corresponding RCPFs. Its three-stage framework integrates unconstrained global exploration, event-driven bidirectional coevolution, and final convergence refinement. Experiments on 87 benchmark instances from seven test suites and 28 real-world engineering CMOPs demonstrate that DCF2D achieves the best overall performance among nine algorithms.  Code available at: https://github.com/RuiqingS/DCF2D.
\end{abstract}

	\begin{IEEEkeywords}
		Constraint handling, Coevolutionary algorithm, Constraint decoupling.
	\end{IEEEkeywords}
\section{Introduction}
\label{c1}

Many engineering and scientific applications involve optimizing multiple conflicting objectives under multiple constraints \cite{zx1,zx2,momo}. Such problems are known as constrained multi-objective optimization problems (CMOPs). A CMOP can be formulated as follows \cite{CCMO}:
\begin{equation}
    \label{CMOPs}
    \left\{
    \begin{aligned}
        & \min \; F(X)=(f_1(X),\ldots,f_M(X)), \\
        & \text{subject to} \quad X \in \Omega,\\
        & g_i(X) \leq 0, \quad i=1,\ldots,p, \\
        & h_j(X) = 0, \quad j=1,\ldots,q,
    \end{aligned}
    \right.
\end{equation}
where $X\in\Omega\subseteq\mathbb{R}^D$ is a decision vector and $F(X)$ contains $M$ objective functions. The total number of constraints is $n_{con}=p+q$.

The violation of the $k$-th constraint is defined as \cite{nsga2}
\begin{equation}
    \label{cjx}
    c_k(X)=
    \left\{
    \begin{aligned}
        & \max(0,g_k(X)), && k=1,\ldots,p, \\
        & \max(0,|h_{k-p}(X)|-\tau),
        && k=p+1,\ldots,n_{con},
    \end{aligned}
    \right.
\end{equation}
where $\tau$ is a small tolerance for equality constraints. The total constraint violation is
\begin{equation}
    \label{CVX}
    CV(X)=\sum_{k=1}^{n_{con}}c_k(X).
\end{equation}
A solution is feasible if $CV(X)=0$. Let
\begin{equation}
    \mathcal{F}=\{X\in\Omega\mid CV(X)=0\}.
\end{equation}
Using $\prec$ to denote Pareto dominance, the constrained Pareto front (CPF) and unconstrained Pareto front (UPF) are respectively defined as
\begin{equation}
    \mathrm{CPF}
    =
    \{F(X)\mid X\in\mathcal{F},
    \nexists\,Y\in\mathcal{F}:F(Y)\prec F(X)\},
\end{equation}
and
\begin{equation}
    \mathrm{UPF}
    =
    \{F(X)\mid X\in\Omega,
    \nexists\,Y\in\Omega:F(Y)\prec F(X)\}.
\end{equation}

Constrained multi-objective evolutionary algorithms (CMOEAs) combine multi-objective evolutionary algorithms with constraint-handling techniques (CHTs) to balance feasibility, convergence, and diversity \cite{CCMO,ctaea,rlsd}. Most CHTs aggregate all constraint violations into a scalar value in Eq.~(\ref{CVX}). This treatment ignores the distinct roles and coupling relationships of individual constraints. It may therefore impede optimization when only some constraints shape the final CPF.

Constraint-decoupling methods such as MSCMO \cite{MSCMO}, C3M \cite{c3m}, MCCMO \cite{mccmo}, and MTOTC \cite{mtotc} address this issue by handling constraints separately. However, two limitations remain. First, most methods decouple all constraints even when some are irrelevant to the CPF. This creates unnecessary computational overhead. Second, they generally search only in the evolutionary direction to approximate single-constraint Pareto fronts (SCPFs). Such a search may fail when the CPF is determined by an infeasible boundary rather than by any SCPF.

For $X,Y\in\Omega$, the direction from $X$ to $Y$ is called the \textit{evolutionary direction} if $F(Y)\prec F(X)$. It is called the \textit{reverse direction} if $F(X)\prec F(Y)$. The former searches toward nondominated regions. The latter intentionally searches toward dominated regions.

To analyze the role of each constraint, let
\begin{equation}
    \mathcal{F}_i=\{X\in\Omega\mid c_i(X)=0\}.
\end{equation}
The single-constraint Pareto front associated with the $i$-th constraint is
\begin{equation}
    \label{eq:scpf}
    \mathrm{SCPF}_i
    =
    \{F(X)\mid X\in\mathcal{F}_i,
    \nexists\,Y\in\mathcal{F}_i:F(Y)\prec F(X)\}.
\end{equation}
We define the portion of the CPF outside all SCPFs as the independent constrained Pareto front (ICPF):
\begin{equation}
    \mathrm{ICPF}
    =
    \mathrm{CPF}
    \setminus
    \bigcup_{i=1}^{n_{con}}\mathrm{SCPF}_i.
\end{equation}
Intuitively, an ICPF point becomes Pareto-optimal only through the joint effect of multiple constraints. Under any individual constraint, it is dominated by a solution satisfying that constraint; however, each such dominating solution violates at least one of the other constraints. Thus, the ICPF arises from constraint coupling rather than from any single constraint.

To provide guidance for discovering the ICPF, define the infeasible region of the $i$-th constraint as
\begin{equation}
    \mathcal{G}_i=\{X\in\Omega\mid c_i(X)>0\}.
\end{equation}
The associated candidate set is
\begin{equation}
    \mathcal{B}_i
    =
    \{F(X)\mid X\in\mathcal{G}_i,
    \exists\, \mathbf{y}\in\mathrm{ICPF}: F(X)\prec \mathbf{y}\}.
\end{equation}
The reverse constrained Pareto front (RCPF) is then defined as
\begin{equation}
    \mathrm{RCPF}_i
    =
    \{u\in\mathcal{B}_i\mid
    \nexists\,v\in\mathcal{B}_i:u\prec v\}.
\end{equation}
The RCPF is the reverse nondominated boundary of infeasible solutions that dominate the ICPF. It provides reverse-direction guidance for discovering the ICPF. If the ICPF is empty, the associated $\mathcal{B}_i$ and $\mathrm{RCPF}_i$ are also empty.

\begin{figure*}[htbp]
    \centering
    \subfigure[]{
        \includegraphics[width=4cm]{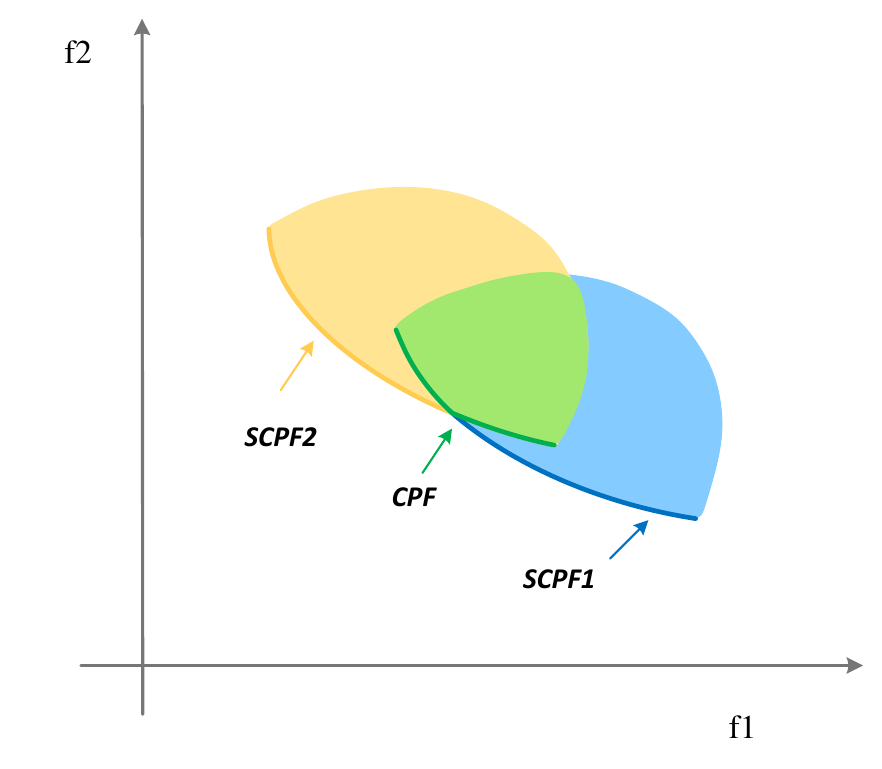}
        \label{type1}
    }
    \subfigure[]{
        \includegraphics[width=4cm]{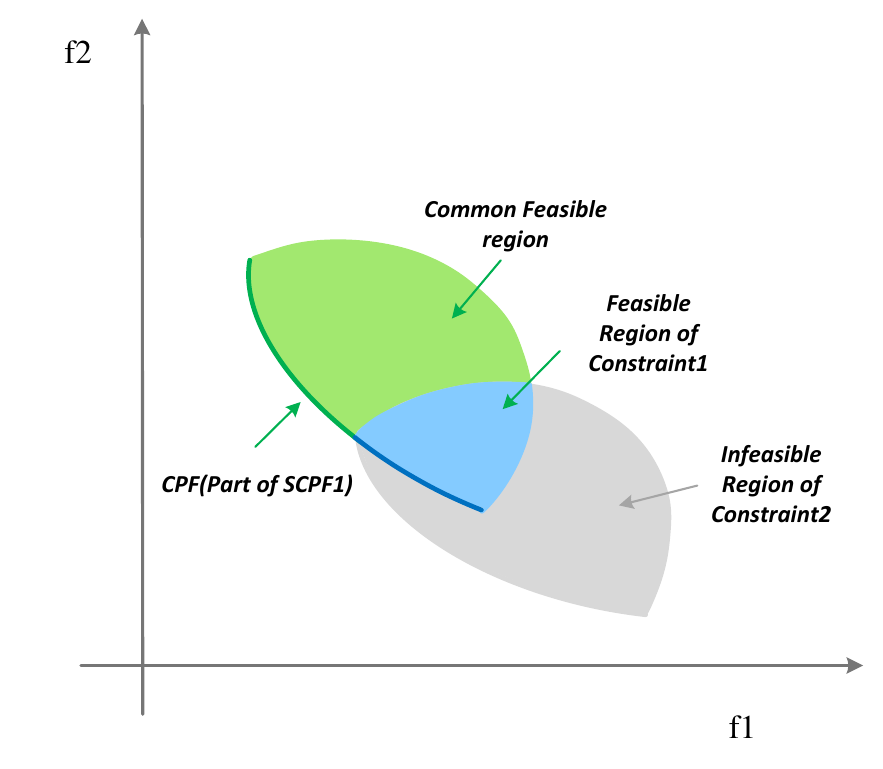}
        \label{type2}
    }
    \subfigure[]{
        \includegraphics[width=4cm]{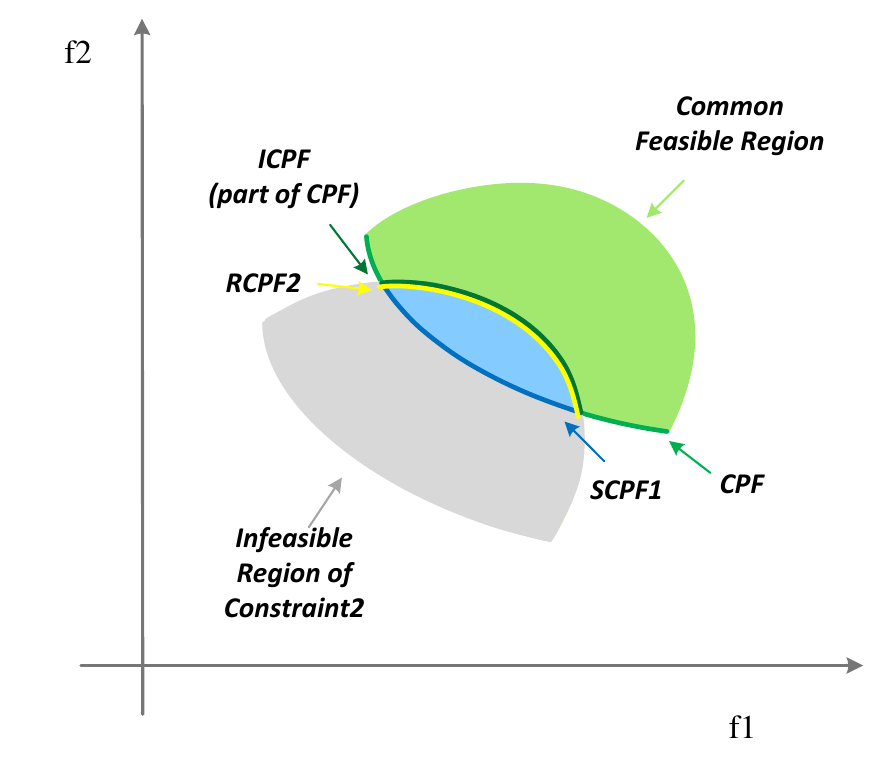}
        \label{type3}
    }
    \subfigure[]{
        \includegraphics[width=4cm]{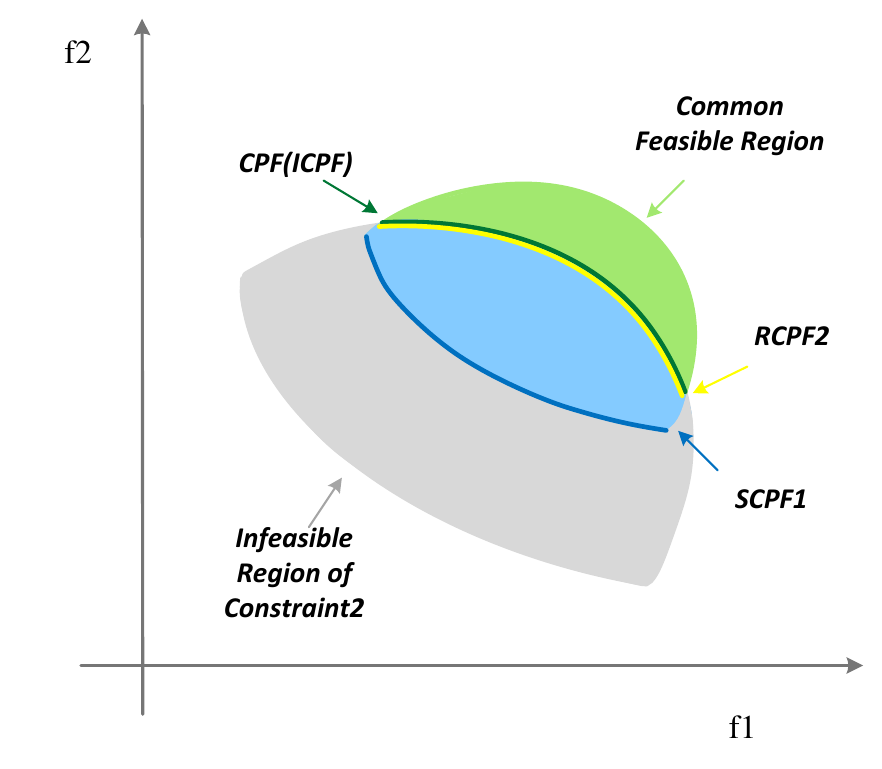}
        \label{type4}
    }
    \caption{Four types of constraint coupling:
    (a) the CPF consists of segments from two SCPFs;
    (b) the CPF lies on one SCPF and is bounded by another constraint;
    (c) the CPF contains both SCPF and ICPF segments;
    and (d) the CPF consists entirely of the ICPF.}
    \label{type}
\end{figure*}

Fig.~\ref{type} illustrates four representative coupling cases. In Fig.~\ref{type1}, two SCPFs jointly form the CPF. In Fig.~\ref{type2}, only one SCPF contributes to the CPF. These cases can be addressed by searching along the evolutionary direction. In Fig.~\ref{type3}, the CPF contains both SCPF and ICPF segments. The SCPF requires evolutionary-direction search, whereas the ICPF requires reverse-direction guidance from the RCPF. In Fig.~\ref{type4}, the CPF consists entirely of the ICPF. Approximating the RCPF is therefore essential. These cases show that effective constraint decoupling must identify relevant constraints and search in both directions.

Based on these observations, we propose DCF2D (Decoupling Constraints from Two Directions). DCF2D operates in three adaptive stages. First, an auxiliary population ignores all constraints and explores the UPF. It is deactivated after the search stabilizes. Second, an event-driven mechanism identifies the constraints that block the main population. DCF2D then creates auxiliary populations for these constraints and assigns each population an evolutionary or reverse search direction. Third, all auxiliary populations are deactivated. The computational resources are transferred to the main population for final convergence.

The main contributions are summarized as follows:
\begin{enumerate}
    \item We analyze how multiple constraints jointly shape the CPF. Based on this analysis, we introduce the ICPF and RCPF to characterize CPF segments that cannot be discovered through SCPFs alone.
    
    \item We develop a dynamic mechanism that identifies active constraints and determines the useful search direction for each constraint. This mechanism avoids unnecessary constraint decoupling.
    
    \item We propose DCF2D to exploit constraint coupling in both the evolutionary and reverse directions. Experimental results demonstrate its effectiveness on complex CMOPs with multiple constraints.
\end{enumerate}

The remainder of this paper is organized as follows. Section~II reviews related work. Section~III presents DCF2D. Section~IV reports the experimental results. Section~V concludes the paper.

\section{Related Work}

\begin{table*}[b]
  \centering
  \caption{Feature comparison between DCF2D and representative state-of-the-art CMOEAs}
  \resizebox{\textwidth}{!}{
  \begin{tabular}{lcccccc}
    \toprule
    \textbf{Algorithm} & \textbf{Year} & \textbf{Category} & \textbf{Constraint Handling} & \textbf{Decoupling Strategy} & \textbf{Search Direction} & \textbf{Key Mechanism} \\
    \midrule
    MSCMO \cite{MSCMO} & 2021 & Decoupling & Progressive & Sequential & Unidirectional & Infeasibility Rate Priority\\
    C3M \cite{c3m} & 2022 & Decoupling & Classification & Sequential & Unidirectional & Pareto Priority \\
    MCCMO \cite{mccmo} & 2023 & Decoupling & Auto Sub-problems & Merge & Unidirectional & Dynamic Population Merging \\
    MTOTC \cite{mtotc} & 2024 & Decoupling & Constraint Ignoring & One-off & Unidirectional & Task Cloning \\
    FCDS \cite{fcds} & 2025 & Decoupling & Fuzzy Relaxation & Level-based & Unidirectional & Discrete Satisfaction Levels \\
    \textbf{DCF2D (Proposed)} & \textbf{2026} & \textbf{Decoupling} & \textbf{Bidirectional (SCPF \& RCPF)} & \textbf{Event-driven} & \textbf{Bidirectional} & \textbf{Adaptive Direction Switching} \\
    \bottomrule
  \end{tabular}
  }
  \label{tab:comparison}
\end{table*}
Existing CMOEAs can be broadly classified into traditional constraint-handling, coevolutionary, multi-stage, machine-learning-assisted, and constraint-decoupling methods. The first four categories mainly improve the balance among feasibility, convergence, and diversity, whereas constraint-decoupling methods further distinguish the roles of individual constraints.

\subsection{Traditional Constraint-Handling Methods}

Traditional constraint-handling techniques generally aggregate multiple constraint violations into the overall constraint violation $CV$ and use it to modify solution comparison or objective evaluation. The constraint dominance principle (CDP) \cite{nsga2} prioritizes feasible solutions over infeasible ones and compares two infeasible solutions according to their overall constraint violations. The $\epsilon$-constrained method temporarily treats solutions with violations below a dynamic threshold as feasible, thereby relaxing feasibility pressure during early evolution. Penalty-function methods \cite{spf1} incorporate $CV(X)$ into the objective functions so that objective quality and constraint satisfaction can be optimized within a unified fitness measure. Stochastic ranking \cite{sr} probabilistically alternates between objective-based and penalty-based comparisons to avoid excessive feasibility pressure. Multi-objective-based frameworks \cite{mob} treat objective optimization and constraint satisfaction as separate optimization criteria. 

These early CHTs are typically applied to a single population based on the overall constraint violation $CV$. Although they ignore the relationships among individual constraints, they remain effective for problems with relatively simple constraint structures. Rather than completely replacing them, most subsequent CMOEAs incorporate their underlying principles as essential components within more sophisticated evolutionary frameworks.

\subsection{Coevolutionary and Multi-population Methods}

To overcome the limitations of single-population search, coevolutionary and multi-population methods employ multiple populations, tasks, or archives with complementary roles. These methods mainly follow several ideas. The first is to coordinate constrained and unconstrained searches, allowing auxiliary populations to explore regions that are difficult to reach under strict feasibility pressure; representative methods include CCMO \cite{CCMO}, BiCo \cite{bico}, Dp-ACS \cite{dpacs}, and cDPEA \cite{cdpea}. The second is to assign different populations or archives complementary roles in convergence, feasibility, and diversity preservation, as adopted in C-TAEA \cite{ctaea}, CMOEA-TCP \cite{cmoeatcp}, DBC-CMOEA \cite{dbccmoea}, TPEA \cite{tpea}, and CMOEA-CD \cite{cmoeacd}. The third is to enhance offspring generation and information transfer among populations or tasks, including EMCMO \cite{emcmo}, CMOSMA \cite{cmosma}, CCMODE \cite{ccmode}, EMCMMS \cite{emcmms}, and DBEMTO \cite{dbemto}. Another idea is to adaptively manage population resources or constraint pressure during collaboration, represented by ACREA \cite{acrea}, APSEA \cite{apsea}, MTCMO \cite{mtcmo}, and TriP \cite{trip}.

Overall, these methods exploit complementary populations to cross infeasible barriers and balance feasibility, convergence, and diversity. However, most of them still incorporate conventional CHTs based on the overall constraint violation $CV$ and thus do not explicitly characterize the roles or coupling relationships of individual constraints.

\subsection{Multi-stage Methods}

Multi-stage methods divide evolution into phases with different optimization preferences. These methods generally follow three ideas. The first is to separate objective-space exploration from feasible convergence, with representative methods including ToP \cite{top}, TSTI \cite{tsti}, TSCSO \cite{tscso}, and CMOES \cite{cmoes}. The second is to adjust constraint pressure or the relative importance of objectives and constraints according to the population state, as in MOEA/D-DAE \cite{dae}, CMOEA\_MS \cite{cmoeams}, and MOEA/D-LCDP \cite{lcdp}. The third is to introduce transitional stages or progressively tightened intermediate problems to connect relaxed exploration with the original CMOP, represented by CMOEMT \cite{cmoemt}, CAEAD \cite{caead}, MSCEA \cite{mscea}, and DD-CMOEA \cite{ddcmoea}.

Overall, multi-stage methods alleviate the conflict between objective optimization and constraint satisfaction by varying the search preference over time. Nevertheless, they often rely on predefined stages, thresholds, or switching criteria and generally process the complete constraint set or aggregated $CV$, without explicitly modeling the interactions among individual constraints.

\subsection{Machine-Learning-Assisted Methods}

Machine-learning-assisted methods learn when to activate search strategies or how to configure algorithmic components from population states. RLSD \cite{rlsd} uses reinforcement learning to determine the current evolutionary stage and select the corresponding search strategy. CMOQLMT \cite{cmoqlmt} applies Q-learning to choose an auxiliary optimization task according to feedback from the evolving population. CMOEA-AOP \cite{shao2026deep} uses a deep deterministic policy gradient to adaptively determine the combination ratios of multiple variation operators. DRL-SAEA \cite{shao2025deep} employs Double DQN to select surrogate-modeling schemes for computationally expensive CMOPs. RLANS \cite{li2026reinforcement} uses reinforcement learning to switch among niche-preservation strategies for exploring disconnected feasible regions. GCEA-DSC \cite{wu2026group} applies DQN to alternate between objective-space and decision-space clustering according to their current search effectiveness. MetaMTO \cite{zhan2026learning} learns where, what, and how to transfer knowledge among optimization tasks. Other methods like DD-CMOEA \cite{ddcmoea}, DVCEA \cite{dvcea}, and CG-PPS \cite{cgpps} employ adaptive policies, classification, or clustering techniques to assist optimization.

These methods improve adaptation by learning search decisions online or offline, but they may introduce training and inference costs. More importantly, they usually regard the aggregated $CV$ as a state feature and learn constraint-handling policies implicitly, rather than explicitly identifying the geometric roles and coupling relationships of individual constraints.

\subsection{Constraint-Decoupling Methods}

Constraint-decoupling methods process individual constraints or constraint subsets separately, thereby exposing information hidden by the aggregated $CV$. MSCMO \cite{MSCMO} ranks constraints according to their infeasibility rates and progressively incorporates them into the search, so that constraints with different levels of difficulty can be handled sequentially. C3M \cite{c3m} classifies constraints according to their relationships with the SCPF and prioritizes those that are more relevant to the final CPF. MCCMO \cite{mccmo} initially assigns constraints to separate subpopulations and dynamically merges subproblems when dominance relationships indicate that their constraints should be optimized jointly. MTOTC \cite{mtotc} clones auxiliary tasks that ignore selected constraints, allowing the resulting relaxed subproblems to provide search information to the original CMOP. FCDS \cite{fcds} avoids directly comparing the magnitudes of constraint violations. Instead, it evaluates a solution according to the number of constraints it satisfies, and maps this count to discrete fuzzy satisfaction levels.

However, existing decoupling methods primarily search in the evolutionary direction. DCF2D addresses this by decoupling constraints in both evolution (SCPF) and reverse (RCPF) directions based on explicit coupling detection.

\section{PROPOSED ALGORITHM}
\label{sec:algorithm}

\subsection{Framework of DCF2D}
\label{sec:framework}

DCF2D consists of three adaptive stages: global exploration, co-evolution with constraint decoupling, and final convergence. The overall procedure is presented in Algorithm~\ref{alg:dcf2d}. The key variables are defined as follows:
$MP$: the main population, consisting of $N$ individuals and responsible for approximating the CPF.    
$AP_i$: the $i$-th auxiliary population. Specifically, $AP_0$ contains $N$ individuals and explores the UPF, whereas each constraint-specific population $AP_i$, $i=1,\ldots,n_{con}$, contains $N/4$ individuals and handles the $i$-th constraint.
$LI$: a boundary archive that stores informative infeasible solutions and, when necessary, feasible boundary anchors.
$D_i$: the search direction of $AP_i$. $True$ denotes the \textit{evolutionary} direction for approximating the SCPF, whereas $False$ denotes the \textit{reverse} direction for approximating the RCPF.
$Act_i$: a Boolean flag indicating whether $AP_i$ is active.

\begin{algorithm}[h]
    \small
    \caption{Framework of DCF2D}
    \label{alg:dcf2d}
    
    \KwIn{$N$ (population size), $\beta$ (Stage 3 activation threshold)}
    \KwOut{$MP$ (final solutions)}
    
    \textbf{Initialize:} $n_{con} \gets$ constraint count; $MP \gets$ Init($N$); $AP_{0 \dots n_{con}} \gets \emptyset$\;
    \textbf{State:} $Stage \gets 1$; $Act_{0} \gets True$; $Act_{1 \dots n_{con}} \gets False$; $D_{1 \dots n_{con}} \gets Evolutionary$\;
    $AP_0 \gets MP$; $LI \gets \emptyset$\;
    
    \While{the termination criterion is not met}{
        
        \tcp{Reproduction (Dynamic Resource Allocation)}
        Allocate offspring budget: 50\% to $MP$, 50\% shared among active $AP_i$\;
        Generate offspring set $\mathcal{O}$ using DE operators\;
        
        \tcp{Update Populations}
        Update $MP$ using $\mathcal{O}$ based on CDP\;
        Update $AP_0$ using $\mathcal{O}$ based on raw objective values\;
        Update active $AP_{1 \dots n_{con}}$ using $\mathcal{O}$ in direction $D_i$ (Alg. \ref{alg:updateAP})\; 
        Update $LI$ using $\mathcal{O}$ (Alg. \ref{alg:updateLI})\;
        
        \tcp{Global Phase Transition to Stage 3}
        \If{$FE > \beta \times MaxFE$ \textbf{and} $Stage \neq 3$}{
            Deactivate all $AP_i$\;
            $Stage \gets 3$\;
        }
        \tcp{Event-Driven Activation}
        \If{$LI$ is updated \textbf{and} $Stage = 2$}{
            $infCon \gets$ Detect violated constraints in new $LI$ solutions\;
            \ForEach{$j \in infCon$}{
                \If{$AP_j$ is inactive}{
                    $Act_j \gets True$; $D_j \gets Evolutionary$; Set Protection for $AP_j$\;
                }
            }
        }
        
        \Switch{$Stage$}{
            \Case{1 (Exploration)}{
                Calculate Centroid Movement $\Delta M_t$ of $AP_0$\;
                \If{Eq. \ref{eq:move} is satisfied}{
                    $Stage \gets 2$; $Act_0 \gets False$; \tcp{Stop Global Exploration}
                    $infCon \gets$ Detect violated constraints in $LI$\;
                    Activate $AP_i$ for all $i \in infCon$\;
                }
            }
            \Case{2 (Co-evolution with Decoupling)}{
                \ForEach{active $AP_i$ ($i \in \{1 \dots n_{con}\}$)}{
                    \If{$D_i$ is $Evolutionary$ \textbf{and} no feasible solution in $AP_i$}{
                            $D_i \gets Reverse$; \tcp{Switch to RCPF search}
                    }
                    \ElseIf{$D_i$ is $Reverse$ \textbf{and} $AP_i$ is dominated by $MP$ \textbf{and} not Protected}{
                            $Act_i \gets False$; \tcp{Deactivate}
                    }
                }
            }
            \Case{3 (Convergence)}{
                \tcp{Pure exploitation}
                Continue evolving $MP$\;
            }
        }
    }
    \Return $MP$\;
\end{algorithm}
\textbf{1) Initialization and Reproduction:}
Initially, $MP$ and $AP_0$ are randomly initialized. DCF2D employs a dynamic resource allocation strategy to balance computational efficiency and search diversity. In each generation, the total offspring budget is fixed at $N$. Half of the budget, i.e., $N/2$, is assigned to $MP$ to maintain convergence pressure, while the remaining half is equally distributed among the active auxiliary populations.

Each constraint-specific auxiliary population has a target size of $N/4$. This reduced size is adopted for two reasons. First, environmental selection commonly has quadratic complexity with respect to population size, and maintaining full-sized auxiliary populations would substantially increase the computational cost. Second, the auxiliary populations are intended to locate informative constraint boundaries rather than precisely approximate the final CPF. They therefore require less search precision than $MP$. Reducing their sizes improves efficiency while retaining sufficient boundary-tracking capability. To support the Differential Evolution (DE) operators, each active auxiliary population is assigned a minimum offspring quota; this quota is set to five in the experiments.

DCF2D employs two complementary DE operators. DE/rand/1 provides strong global exploration and facilitates the discovery of disconnected feasible regions induced by complex constraints. In contrast, DE/current-to-pbest/1 introduces directional pressure toward superior individuals. This property enables the auxiliary populations to approach and track the corresponding SCPFs or RCPFs efficiently, thereby providing informative guidance to $MP$.

\textbf{2) Stage 1: Global Exploration via Trajectory Analysis:}
In Stage 1, $AP_0$ ignores all constraints and approximates the UPF. A sliding-window trajectory analysis is used to detect its convergence while reducing premature stage transitions caused by evolutionary fluctuations. Let $\mathcal{C}_t$ denote the centroid of $AP_0$ in the objective space at generation $t$. Its relative movement is defined as
\begin{equation}
    M_t =
    \frac{\|\mathcal{C}_t-\mathcal{C}_{t-1}\|}
    {\|\mathcal{C}_t\|+\tau},
\end{equation}
where $\tau$ is a small positive constant that prevents division by zero. A sliding window of size $W=5$ records the recent movements. The algorithm enters Stage 2 when their average falls below a threshold $\delta$:
\begin{equation}
    \label{eq:move}
    \frac{1}{W}\sum_{k=0}^{W-1}M_{t-k}<\delta,
    \qquad \delta=10^{-4}.
\end{equation}
After the transition, $AP_0$ is deactivated to release computational resources. The algorithm then examines $LI$ and activates the auxiliary populations associated with the detected blocking constraints.

\textbf{3) Stage 2: Event-Driven Co-evolution:}
Stage 2 dynamically decouples the constraints through an event-driven mechanism consisting of population activation, direction switching, and population deactivation.

\textbf{Activation:}
Whenever Algorithm~\ref{alg:updateLI} adds new boundary solutions to $LI$, DCF2D identifies the constraints violated by these solutions. If constraint $j$ is violated and $AP_j$ is inactive, $AP_j$ is activated and assigned a protection period (5 generations). Consequently, auxiliary populations are created only for constraints that currently obstruct the convergence of $MP$.
    
 \textbf{Direction switching and deactivation:}
Each active auxiliary population is continuously monitored. An $AP_i$ initially searches in the \textit{evolutionary} direction to approximate $\mathrm{SCPF}_i$. If it contains no solution satisfying constraint $i$, its direction is switched to \textit{reverse} to approximate $\mathrm{RCPF}_i$. Conversely, because RCPF solutions should dominate the CPF, a reverse-direction $AP_i$ is deactivated if all its solutions become dominated by $MP$ after the protection period expires. This condition indicates that the corresponding infeasible barrier no longer provides useful guidance.

This event-driven strategy activates only currently relevant constraints and removes auxiliary populations that no longer contribute to the search, thereby reducing unnecessary computational overhead.

\textbf{4) Stage 3: Final Convergence:}
When the number of function evaluations exceeds $\beta \times MaxFE$, all auxiliary populations are permanently deactivated. The remaining computational resources are allocated to $MP$ to refine the convergence accuracy and distribution of the final feasible solutions.

\begin{figure*}[htbp]
    \centering
    \subfigure[]{
        \includegraphics[width=4cm]{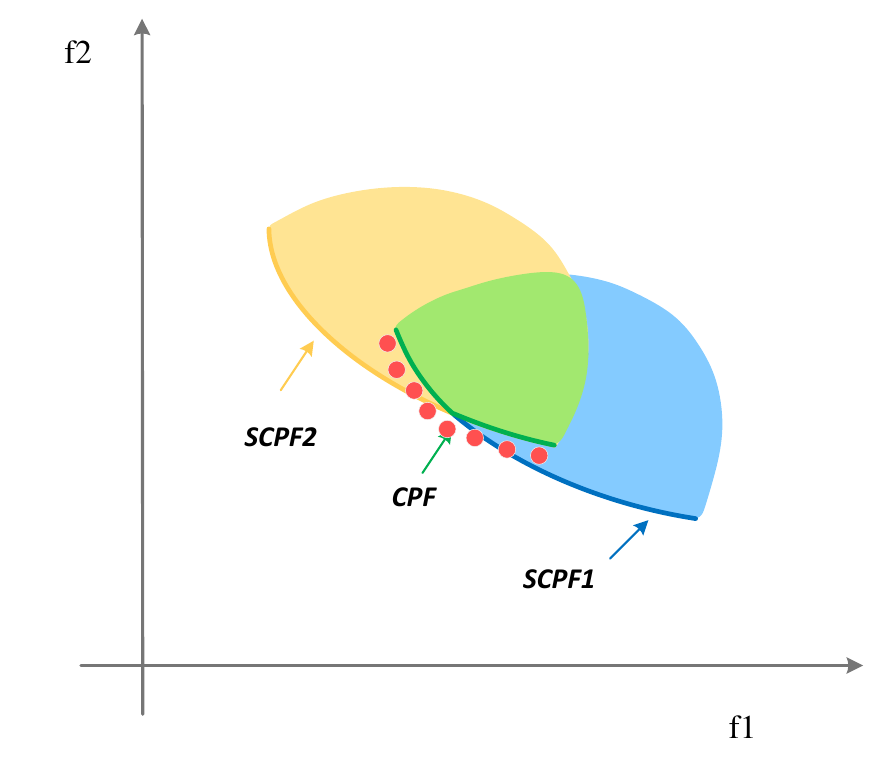}
        \label{type1LI}
    }
    \subfigure[]{
        \includegraphics[width=4cm]{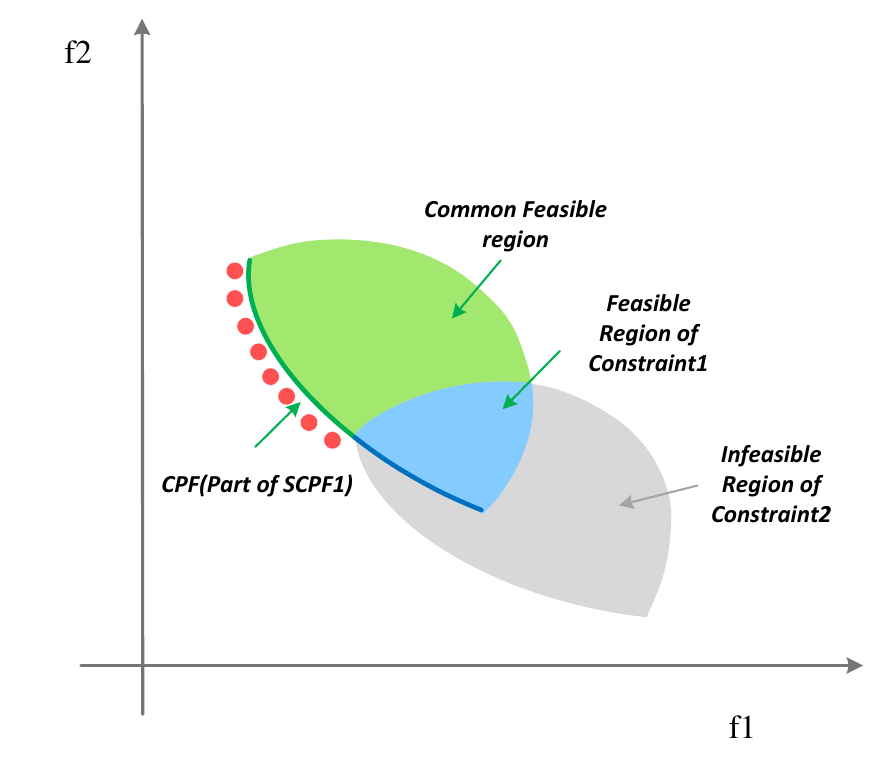}
        \label{type2LI}
    }
    \subfigure[]{
        \includegraphics[width=4cm]{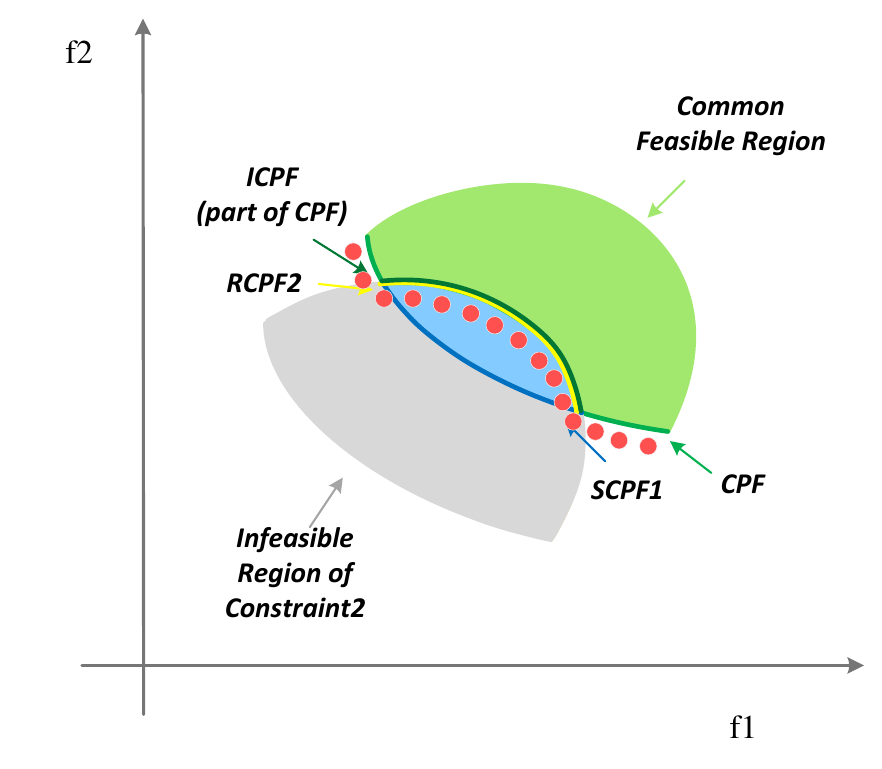}
        \label{type3LI}
    }
    \subfigure[]{
        \includegraphics[width=4cm]{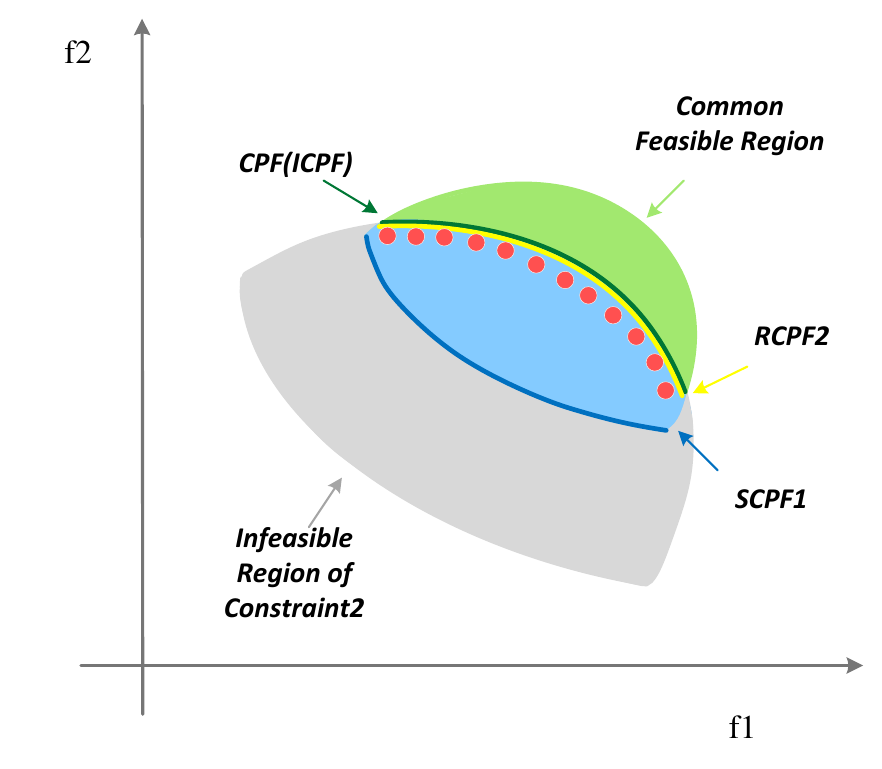}
        \label{type4LI}
    }
    \caption{Distributions of the solutions in the infeasible archive $LI$ under the four constraint-coupling cases. The solutions in $LI$ are marked by red circles. The archive maintains objectively nondominated infeasible solutions discovered during evolution and uses them to identify the constraints that obstruct the convergence of $MP$. Approximating the SCPFs or infeasible boundaries associated with these constraints provides useful directional guidance for DCF2D.}
    \label{typeLI}
\end{figure*}

\subsection{Key functions in DCF2D}

\begin{algorithm}[htbp]
    \small
    \caption{Update Infeasible Archive ($LI$)}
    \label{alg:updateLI}
    \KwIn{$AP_i$ (current auxiliary population), $\mathcal{O}$ (offspring set), 
$N$ (main population size), $i$ (constraint index), $D_i$ (search direction), $MP$ (main population)}
    \KwOut{$LI$ (Updated Archive)}
    
    $S \gets LI \cup \mathcal{O}$\;
    \tcp{Filter 1: Objective-Dominance Screening}
    \tcp{Retain solutions objectively superior to MP}
    $S \gets \{ s \in S \mid (\exists x \in MP: s \prec_{obj} x) \land (\not\exists y \in MP: y \prec_{obj} s) \}$\;
    
    $S_{infeas} \gets$ Select all infeasible solutions in $S$\;
    $S_{feas} \gets$ Select all feasible solutions in $S$\;
    
    \tcp{Filter 2: Boundary-Focused Selection}
    \If{$|S_{infeas}| > N$}{
        \tcp{Prioritize solutions closer to the CPF boundary}
        Calculate fitness of $S_{infeas}$ based on \textbf{negative} objective values ($-F(x)$)\;
        $LI \gets$ Select top $N$ solutions from $S_{infeas}$ based on the calculated fitness\;
    }
    \Else{
        $LI \gets S_{infeas}$\;
        $k \gets N - |S_{infeas}|$\;
        \tcp{Fill remaining slots with best feasible solutions}
        Calculate fitness of $S_{feas}$ based on \textbf{normal} objective values ($F(x)$)\;
        $LI \gets LI \cup$ Select top $k$ solutions from $S_{feas}$\;
    }
    \Return $LI$\;
\end{algorithm}

\subsubsection{Update $LI$}

Algorithm~\ref{alg:updateLI} presents the update procedure for the infeasible archive $LI$. In the event-driven activation mechanism, $LI$ identifies constraints that currently obstruct the convergence of $MP$. To avoid unnecessary activations caused by early evolutionary noise, this mechanism uses $LI$ only from Stage 2 onward. Specifically, $LI$ maintains carefully selected solutions that dominate at least one member of $MP$ and are not dominated by any member of $MP$, thereby capturing informative boundaries associated with the SCPFs or RCPFs. Its update consists of two steps.

\textbf{Step 1: Objective-Dominance Screening.}
The current archive is first merged with the offspring. A solution $s$ is retained only if it objectively dominates at least one solution $x\in MP$ ($s\prec_{obj}x$) and is not objectively dominated by any solution $y\in MP$ ($y\not\prec_{obj}s$). This screening removes inferior solutions caused by evolutionary noise and preserves candidates that are objectively superior to the current main population. The infeasible candidates among them indicate active constraint barriers that may prevent $MP$ from further convergence.

\textbf{Step 2: Boundary-Focused Selection.}
The selection strategy is determined by the number of infeasible candidates:
\begin{itemize}
    \item \textit{Case A: Surplus of infeasible solutions ($|S_{infeas}|>N$).}
    When the number of infeasible candidates exceeds the archive capacity, fitness is calculated using the negative objective values $-F(x)$. For a minimization problem, this reverse selection favors candidates with relatively worse objective values. Among the infeasible solutions located between the UPF and CPF, such candidates are generally farther from the UPF and closer to the feasible boundary. Therefore, selection based on $-F(x)$ enables $LI$ to track the infeasible side of the CPF.

    \item \textit{Case B: Insufficient infeasible solutions ($|S_{infeas}|\leq N$).}
    When the infeasible candidates cannot fill the archive, all of them are retained, and the remaining slots are filled with feasible solutions selected using the original objective values $F(x)$. Feasible solutions with better objective values tend to approach the CPF from the feasible side. They therefore serve as boundary anchors and help $LI$ maintain a dense and continuous representation of the CPF boundary.
\end{itemize}

Through this two-step procedure, $LI$ captures informative solutions around the CPF boundary and supports the identification of constraints that currently impede the convergence of $MP$.

\subsubsection{Update $AP_i$}

\begin{algorithm}[htbp]
    \small
    \caption{Update Constraint-Specific Population ($AP_i$)}
    \label{alg:updateAP}
    \KwIn{$AP_i$ (Current Archive), $\mathcal{O}$ (Offspring), $N$ (Total Size), $i$ (Constraint Index), $D_i$ (Direction), $MP$ (Main Population)}
    \KwOut{$AP_i$ (Updated Archive), $F_i$ (Fitness)}
    
    $S \gets AP_i \cup \mathcal{O}$\;
    $N_{sub} \gets \lfloor N/4 \rfloor$ 
    
    \tcp{Part 1: Evolutionary Direction (Search for SCPF)}
    \If{$D_i$ is $Evolutionary$}{
        $F_i$ $\gets$ Calculate fitness based on $F(x)$ with CDP under $c_i$ (kNN diversity)\;
        [$AP_i, F_i$] $\gets$ Select the top $N_{sub}$ solutions based on $F_i$\;
        \Return $AP_i, F_i$\;
    }
    
    \tcp{Part 2: Reverse Direction (Search for RCPF)}
    $S_{tar} \gets \{s \in S \mid c_i(s) > 0\}$ \tcp*{Violating solutions}
    $S_{res} \gets \{s \in S \mid c_i(s) = 0\}$ \tcp*{Satisfying solutions}

    \eIf{$|S_{tar}| \le N_{sub}$}{
        \tcp{fill with satisfying ones}
        $F_{tar} \gets$ Calculate fitness of $S_{tar}$ using $\{F(x), CV(x)\}$\;
        $K \gets N_{sub} - |S_{tar}|$\;
        $F_{res} \gets$ Calculate fitness of $S_{res}$ based on $F(x)$ with CDP under $c_i$\;
        $AP_i \gets S_{tar} \cup$ Select the top $K$ from $S_{res}$ based on $F_{res}$\;
    }{
        \tcp{perform targeted selection \& squeeze}
        $F_{tar} \gets$ Calculate fitness of $S_{tar}$ using $\{F(x), CV(x)\}$\;
        
        \tcp{Penalize solutions dominated by MP to deprioritize them}
        $S_{dom} \gets \{s \in S_{tar} \mid \exists m \in MP, m \prec s\}$ \;
        $F_{tar}(s) \gets F_{tar}(s) + \max(F_{tar}), \quad \forall s \in S_{dom}$\;
        
        \tcp{Extract the first nondominated front from penalized fitness}
        $S_{next} \gets \{s \in S_{tar} \mid F_{tar}(s) \text{is assigned Rank 1}\}$\;
        
        \eIf{$|S_{next}| \le N_{sub}$}{
            $AP_i \gets$ Select top $N_{sub}$ from $S_{tar}$ based on penalized $F_{tar}$\;
        }{
            \tcp{Geometric Squeeze via Negative Objectives}
            $F_{neg} \gets$ Calculate fitness of $S_{next}$ based on $-F(x)$ (kNN diversity)\;
            $AP_i \gets$ Select top $N_{sub}$ from $S_{next}$ based on $F_{neg}$\;
        }
    }
    
    \Return $AP_i, F_i$\;
\end{algorithm}

Algorithm~\ref{alg:updateAP} presents the update procedure for each constraint-specific auxiliary population $AP_i$. Because these populations primarily explore constraint boundaries, computational efficiency is prioritized over strict diversity preservation. Instead of the computationally expensive iterative archive truncation used in standard SPEA2 \cite{spea2}, DCF2D directly employs $k$-nearest-neighbor ($k$-NN) density estimation. Although this approach provides a less rigorous diversity guarantee, it maintains sufficient distribution pressure for boundary exploration while substantially reducing computational overhead.

When $D_i$ is set to the \textit{evolutionary} direction, $AP_i$ aims to approximate $\mathrm{SCPF}_i$. The current population is first merged with the offspring, after which the CDP is applied by considering only the violation of constraint $i$. The candidates are then ranked and selected according to fitness values incorporating $k$-NN density. This procedure efficiently drives $AP_i$ toward the feasible Pareto front associated with constraint $i$.

When $D_i$ is set to the \textit{reverse} direction, $AP_i$ aims to approximate $\mathrm{RCPF}_i$ from the infeasible side. The candidate set is therefore divided into solutions that violate constraint $i$ ($S_{tar}$) and those that satisfy it ($S_{res}$). If $|S_{tar}|\leq N_{sub}$, all violating solutions are retained, and the remaining slots are filled with satisfying solutions selected using CDP.

If $|S_{tar}|>N_{sub}$, a hierarchical selection procedure is performed. First, the fitness of each solution in $S_{tar}$ is evaluated using both its objective values $F(x)$ and overall $CV(x)$. Jointly considering these two criteria prevents the population from retaining solutions with favorable objective values but excessively large constraint violations. It therefore encourages $AP_i$ to approach the relevant infeasible boundary rather than drift toward lateral regions.

Second, solutions in $S_{tar}$ that are dominated by $MP$ receive the same large fitness penalty. According to the definition of the RCPF, informative RCPF solutions should dominate the CPF. The first nondominated front $S_{next}$ is then extracted using the penalized fitness. If $|S_{next}|\leq N_{sub}$, the remaining slots are filled according to the penalized fitness $F_{tar}$. This strategy gives priority to solutions that dominate $MP$ while retaining dominated solutions only when necessary.

Finally, if $|S_{next}|>N_{sub}$, a geometric squeezing mechanism is applied. Because the solutions in $S_{next}$ dominate $MP$, their fitness is recalculated using the negative objective values $-F(x)$. Reversing the optimization direction favors solutions closer to $MP$, enabling $AP_i$ to track the constraint boundary more tightly and provide effective guidance toward the ICPF.

\subsection{Analysis of DCF2D on a Running Instance}
\label{sec:analysis}

\begin{figure*}[htbp]
    \centering
    \subfigure[Stage 1: Exploration]{
        \includegraphics[width=4cm]{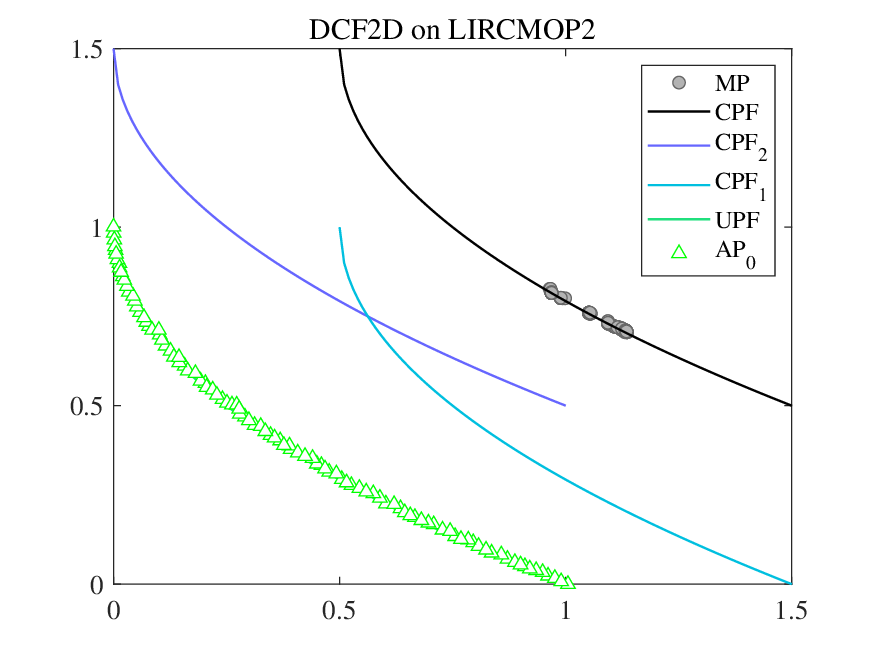}
        \label{lir1}
    }
    \subfigure[Stage 2: Initial Activation]{
        \includegraphics[width=4cm]{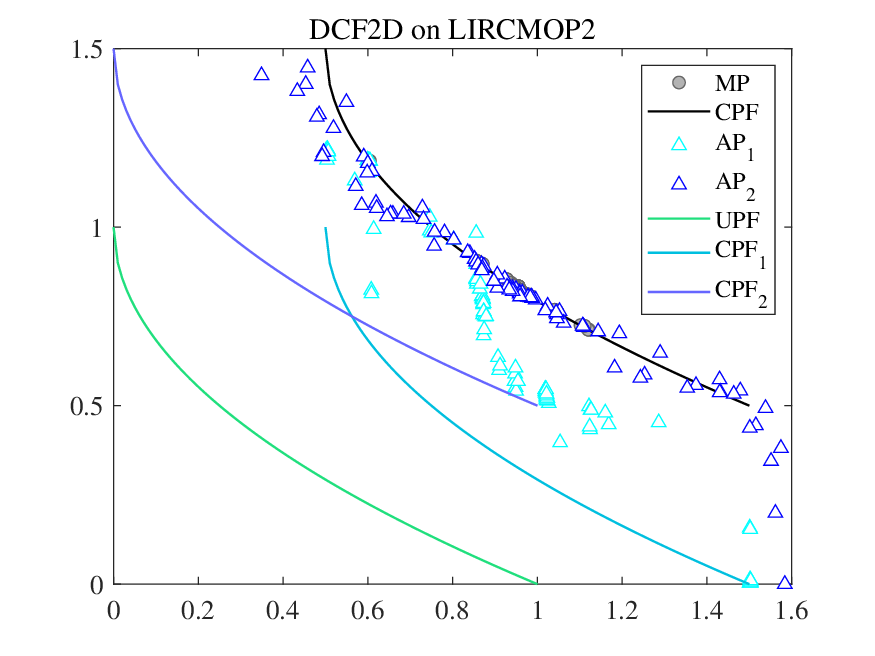}
        \label{lir2}
    }
    \subfigure[Stage 2: Direction Switching]{
        \includegraphics[width=4cm]{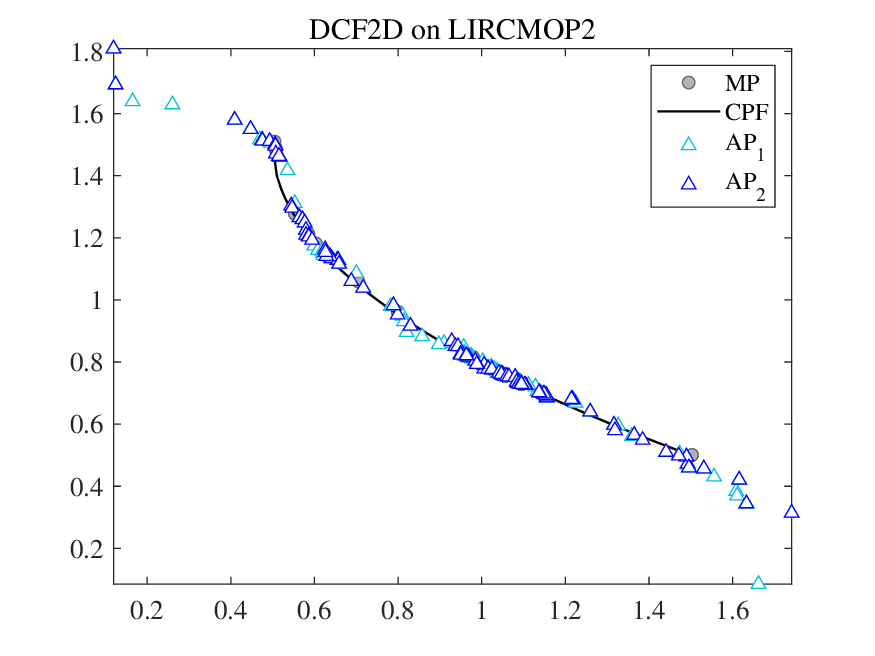}
        \label{lir3}
    }
    \subfigure[Stage 2: RCPF Approximation]{
        \includegraphics[width=4cm]{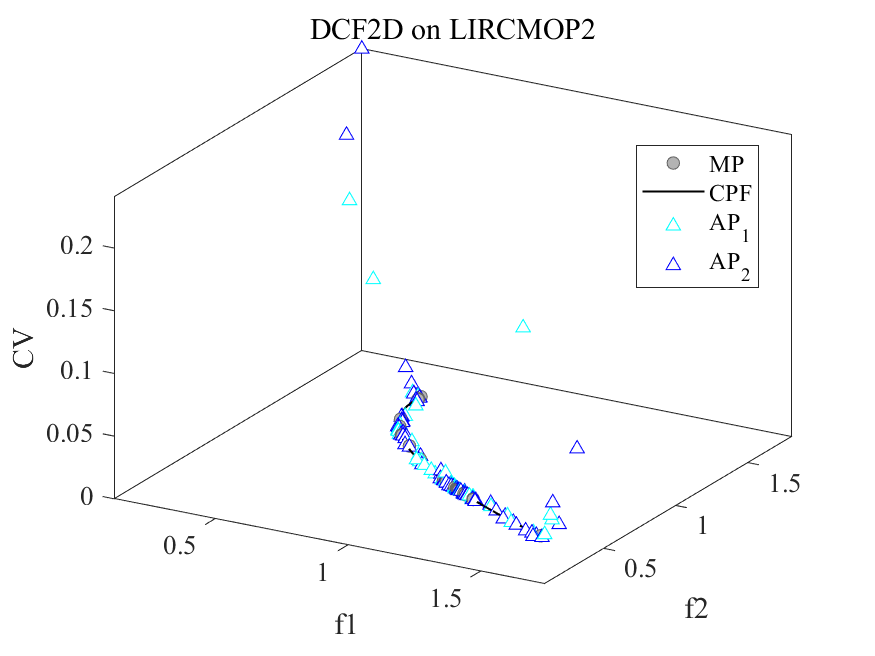}
        \label{lir4}
    }
    \caption{Evolutionary process of DCF2D on LIRCMOP2. (a) In Stage 1, $AP_0$ explores the UPF. (b) At the beginning of Stage 2, $AP_1$ and $AP_2$ are activated in the \textit{evolutionary} direction, with the triangles showing their search toward $\mathrm{SCPF}_1$ and $\mathrm{SCPF}_2$, respectively. (c) Because no feasible solutions are found, both populations switch to the \textit{reverse} direction (inverted triangles). (d) $AP_1$ and $AP_2$ approximate the RCPFs close to the CPF.}
    \label{lir}
\end{figure*}

To illustrate the proposed mechanism, we analyze the evolutionary behavior of DCF2D on LIRCMOP2, a CMOP with two constraints. In Fig.~\ref{lir}, the cyan and purple curves represent $\mathrm{SCPF}_1$ and $\mathrm{SCPF}_2$, respectively, while the green and black curves denote the UPF and CPF. For this problem, the CPF is disjoint from both SCPFs, corresponding to Situation 4 in Section~I (Fig.~\ref{type4}). Consequently, approximating only the SCPFs is insufficient to locate the CPF.
The evolutionary process is summarized as follows:

     \textbf{Stage 1---Exploration (Fig.~\ref{lir1}):}
    $AP_0$ explores the UPF, while $MP$ searches for the CPF. This unconstrained exploration provides information about the location of the UPF relative to the feasible region.

     \textbf{Stage 2---Activation and SCPF search (Fig.~\ref{lir2}):}
    At the beginning of Stage 2, the update procedure of $LI$ in Algorithm~\ref{alg:updateLI} identifies constraints 1 and 2 as current barriers to the convergence of $MP$. Accordingly, $AP_1$ and $AP_2$ are activated in the default \textit{evolutionary} direction, as indicated by the dark-blue triangles, to search for $\mathrm{SCPF}_1$ and $\mathrm{SCPF}_2$.

     \textbf{Stage 2---Switching to RCPF search (Fig.~\ref{lir3}):}
    As evolution proceeds, neither $AP_1$ nor $AP_2$ finds a solution satisfying its corresponding constraint in the \textit{evolutionary} direction because the SCPFs are disconnected from the relevant search region. According to the direction-switching condition in Algorithm~\ref{alg:dcf2d}, $D_1$ and $D_2$ are therefore switched to the \textit{reverse} direction. The two populations subsequently search for $\mathrm{RCPF}_1$ and $\mathrm{RCPF}_2$, as indicated by the light-blue inverted triangles.

     \textbf{Stage 2---RCPF approximation (Fig.~\ref{lir4}):}
    In the \textit{reverse} direction, $AP_1$ and $AP_2$ approximate the corresponding RCPFs from the infeasible side. Following Algorithm~\ref{alg:updateAP}, both objective values and constraint violations are considered during selection. Solutions closer to the relevant constraint boundaries generally have smaller violations, whereas those farther away exhibit larger violations. This selection pressure drives the two auxiliary populations toward the RCPFs close to the CPF.

By combining the boundary information provided by $AP_1$ and $AP_2$, DCF2D guides $MP$ across the infeasible barriers and toward the true CPF. This example demonstrates that dynamically switching from SCPF to RCPF approximation is essential when the CPF is disjoint from the individual SCPFs.

\section{Experimental Studies}
 
\subsection{Experimental Settings}

\subsubsection{Comparative Algorithms and Parameter Settings}
DCF2D is compared with eight state-of-the-art CMOEAs, including five constraint-decoupling methods: C3M \cite{c3m}, MTOTC \cite{mtotc}, MSCMO \cite{MSCMO}, MCCMO \cite{mccmo}, and FCDS \cite{fcds}, and three recent high-performance algorithms: IMTCMO \cite{imtcmo}, ACREA \cite{acrea}, and APSEA \cite{apsea}. The former enable a direct evaluation of the proposed bidirectional decoupling strategy, whereas the latter represents recent advances in handling complex constraints.

Each algorithm uses the reproduction operators and default parameters recommended in its original paper. ACREA and APSEA employ Genetic Algorithm (GA) operators \cite{SBX}; MCCMO uses GA or Differential Evolution (DE) operators depending on the CMOP; and C3M, MTOTC, MSCMO, FCDS, IMTCMO, and DCF2D employ DE operators \cite{DE}. The population size and maximum number of function evaluations (MaxFE) are set to 100 and 300,000, respectively.

Each algorithm is independently run 30 times on each problem using PlatEMO 4.12 \cite{2017PlatEMO} and the original source code provided by the respective authors. The workstation is equipped with an Intel Core i7-13700 CPU, 48 GB RAM, and Windows 11. Statistical significance is assessed using the Wilcoxon rank-sum test at a significance level of 0.05.

\subsubsection{Test CMOPs}
Experiments are conducted on seven benchmark suites: DASCMOP \cite{dascmop}, DOC \cite{top}, LIRCMOP \cite{lir}, MW \cite{mw}, ZXH\_CF \cite{zxhcf}, CDTLZ \cite{cdtlz}, and SDC \cite{imtcmo}. These suites contain multiple constraints and cover all four coupling scenarios discussed in Section I, although multiple scenarios may coexist within one CMOP. To further evaluate DCF2D in practical applications, we also test it on 28 real-world CMOPs (RWMOPs) \cite{rwmop}.

\textbf{DASCMOP} comprises nine difficulty-adjustable CMOPs whose convergence, diversity, and feasibility difficulties are controlled by three parameters.
\textbf{DOC} consists of nine CMOPs with constraints in both objective and decision spaces, resulting in complex constraint-violation landscapes.
\textbf{LIRCMOP} contains fourteen CMOPs characterized by large infeasible regions that are often independent of the CPF but significantly obstruct evolution.
\textbf{MW} comprises fourteen CMOPs covering four relationships between the UPF and CPF: complete overlap, partial overlap, partial separation, and complete separation.
\textbf{ZXH\_CF} includes sixteen problems with two types of constraints: one creates infeasible barriers and guides the search through correlated variables, while the other defines feasible regions with CPFs of various shapes.
\textbf{CDTLZ} introduces multiple inequality constraints that divide the feasible domain into non-convex or multi-segment regions.
\textbf{SDC} directly links constraints to decision variables to approximate real-world characteristics, breaking the linear relationship between objectives and constraints, creating locally optimal infeasible solutions, and incorporating both equality and inequality constraints.
\textbf{RWMOP} comprises 28 real-world CMOPs (RWMOP1--RWMOP29) from the CEC 2021 Real-World Multi-Objective Constrained Optimization Competition. These problems span diverse, highly constrained engineering domains. Specifically, RWMOP1--RWMOP21 concern mechanical design optimization, such as pressure vessel, vibrating platform, and welded beam design; RWMOP22--RWMOP24 arise from chemical engineering, such as reactor and heat-exchanger network design; and RWMOP25--RWMOP29 involve complex process synthesis and design. Unlike synthetic benchmarks, these problems feature strongly coupled physical constraints, highly nonlinear objective spaces, and mixed discrete and continuous variables.

\subsubsection{Performance Indicators}
For the seven benchmark suites with known true CPFs, we use the Modified Inverted Generational Distance (\textbf{IGD+}) \cite{igdm}:
\begin{equation}
    IGD^+(P,Q)=\frac{1}{|P|}\sum_{v\in P}\min_{u\in Q}d^+(v,u),
    \label{IGDplus}
\end{equation}
where $d^+(v,u)=\sqrt{\sum_{i=1}^{M}(\max\{f_i(u)-f_i(v),0\})^2}$, $P$ is a set of reference points uniformly distributed on the true CPF, and $Q$ is the non-dominated solution set obtained by an algorithm. IGD+ is calculated in PlatEMO using 10,000 reference points, with smaller values indicating better performance.

For RWMOPs with unknown true CPFs, we use Hypervolume (HV) \cite{HV}:
\begin{equation}
    HV(Q,R)=L\left(\bigcup_{u\in Q}[f_1(u),r_1]\times\cdots\times[f_M(u),r_M]\right),
    \label{HV}
\end{equation}
where $L(\cdot)$ denotes the Lebesgue measure and $R=(r_1,r_2,\ldots,r_M)$ is the reference point, set to $[1,1,\ldots,1]$ times the nadir point of the combined populations. Larger HV values indicate better performance.

\subsection{Comparison Between DCF2D and State-of-the-Art CMOEAs}

\begin{table*}[htbp]
	\centering
	\caption{Summary of Statistical Significance ($+/-/=$) and Average Runtime (Seconds) on Benchmark Suites and Real-world Problems. `$+$', `$-$', and `$=$' indicate that the comparative algorithm is significantly better than, worse than, or similar to DCF2D, respectively (Wilcoxon rank-sum test, $p < 0.05$). Runtime is measured in seconds.}
	\label{tab:summary}
	\resizebox{\textwidth}{!}{
		\begin{tabular}{ll|ccc|ccccc|c}
			\toprule
			\multirow{2}{*}{\textbf{Suite}} & \multirow{2}{*}{\textbf{Metric}} & \multicolumn{3}{c|}{\textbf{Non-Decoupling SOTA}} & \multicolumn{5}{c|}{\textbf{Constraint Decoupling SOTA}} & \textbf{Proposed} \\
			\cmidrule(lr){3-5} \cmidrule(lr){6-10} \cmidrule(lr){11-11}
			& & ACREA\cite{acrea} & APSEA\cite{apsea} & IMTCMO\cite{imtcmo} & C3M\cite{c3m} & FCDS\cite{fcds} & MCCMO\cite{mccmo} & MSCMO\cite{MSCMO} & MTOTC\cite{mtotc} & DCF2D \\
			\midrule
			
			\multirow{3}{*}{DASCMOP} 
			& Sig. ($+/-/=$) & $0/9/0$ & $1/6/2$ & $0/5/4$ & $0/9/0$ & $0/8/1$ & $0/9/0$ & $0/9/0$ & $0/9/0$ & -- \\
			& Runtime (s) & 91.23 & \textbf{15.77} & 57.54 & 60.30 & 155.54 & 77.87 & 29.71 & 253.92 & 72.51 \\
			& Avg. Rank & 5.33 & 5.78 & 2.22 & 5.78 & 4.33 & 6.11 & 6.89 & 7.22 & \textbf{1.33} \\
			\midrule
			
			\multirow{3}{*}{DOC} 
			& Sig. ($+/-/=$) & $0/9/0$ & $0/9/0$ & $0/3/6$ & $3/5/1$ & $0/8/1$ & $4/5/0$ & $3/5/1$ & $1/3/5$ & -- \\
			& Runtime (s) & 96.73 & \textbf{34.34} & 89.40 & 69.87 & 182.22 & 50.90 & 60.75 & 310.44 & 127.35 \\
			& Avg. Rank & 9.00 & 8.11 & 3.56 & 3.89 & 6.44 & 3.11 & 3.78 & 4.67 & \textbf{2.89} \\
			\midrule
			
			\multirow{3}{*}{CDTLZ} 
			& Sig. ($+/-/=$) & $5/3/2$ & $4/4/2$ & $0/10/0$ & $0/10/0$ & $0/10/0$ & $0/10/0$ & $0/10/0$ & $0/10/0$ & -- \\
			& Runtime (s) & 143.80 & 42.22 & 63.48 & 28.56 & 326.83 & 58.12 & \textbf{28.13} & 143.59 & 96.97 \\
			& Avg. Rank & 2.30 & 3.90 & 5.20 & 6.70 & 3.50 & 6.40 & 8.40 & 6.50 & \textbf{2.10} \\
			\midrule
			
			\multirow{3}{*}{LIRCMOP} 
			& Sig. ($+/-/=$) & $6/8/0$ & $0/14/0$ & $3/7/4$ & $0/13/1$ & $3/7/4$ & $2/11/1$ & $0/12/2$ & $3/8/3$ & -- \\
			& Runtime (s) & 114.02 & \textbf{17.11} & 98.10 & 41.74 & 220.09 & 69.88 & 36.56 & 231.57 & 72.42 \\
			& Avg. Rank & 3.14 & 9.00 & 4.36 & 6.43 & 4.00 & 4.64 & 6.00 & 4.71 & \textbf{2.71} \\
			\midrule
			
			\multirow{3}{*}{MW} 
			& Sig. ($+/-/=$) & $1/11/2$ & $8/5/1$ & $2/9/3$ & $0/14/0$ & $5/6/3$ & $0/14/0$ & $0/14/0$ & $0/13/1$ & -- \\
			& Runtime (s) & 111.83 & 29.02 & 61.14 & 22.04 & 257.73 & 47.87 & \textbf{20.04} & 121.16 & 76.93 \\
			& Avg. Rank & 5.21 & 2.21 & 3.50 & 6.93 & 2.71 & 7.21 & 7.86 & 6.79 & \textbf{2.57} \\
			\midrule
			
			\multirow{3}{*}{SDC} 
			& Sig. ($+/-/=$) & $0/14/1$ & $1/12/2$ & $7/3/5$ & $3/8/4$ & $3/7/5$ & $2/8/5$ & $3/7/5$ & $2/10/3$ & -- \\
			& Runtime (s) & 139.66 & 29.00 & 34.51 & 28.45 & 177.16 & 39.17 & \textbf{28.08} & 91.69 & 63.42 \\
			& Avg. Rank & 8.07 & 5.93 & 2.33 & 5.67 & 4.40 & 4.93 & 4.33 & 6.20 & \textbf{3.20} \\
			\midrule
			
			\multirow{3}{*}{ZXH\_CF} 
			& Sig. ($+/-/=$) & $10/4/2$ & $7/3/6$ & $0/13/3$ & $1/12/3$ & $1/12/3$ & $0/12/4$ & $0/12/4$ & $0/15/1$ & -- \\
			& Runtime (s) & 136.10 & 39.63 & 91.81 & 42.05 & 247.75 & 104.35 & \textbf{39.04} & 427.02 & 87.89 \\
			& Avg. Rank & \textbf{3.06} & 3.56 & 5.00 & 6.13 & 3.63 & 6.63 & 7.13 & 6.31 & 3.56 \\
			\midrule
			
			\multirow{3}{*}{RWMOP} 
			& Sig. ($+/-/=$) & $5/21/2$ & $4/16/8$ & $2/15/11$ & $3/18/7$ & $5/12/11$ & $3/16/9$ & $1/16/11$ & $10/10/8$ & -- \\
			& Runtime (s) & 187.73 & \textbf{65.91} & 195.50 & 106.81 & 376.58 & 237.04 & 82.41 & 518.32 & 108.57 \\
			& Avg. Rank & 6.21 & 5.46 & 4.82 & 5.71 & 4.25 & 5.39 & 6.21 & 3.54 & \textbf{3.46} \\
			\midrule
			\midrule
			
			\multirow{3}{*}{\textbf{Overall}} 
			& \textbf{Total Sig.} & 27/79/9 & 25/69/21 & 14/65/36 & 10/89/16 & 17/70/28 & 11/85/19 & 7/85/23 & 16/78/21 & -- \\
			& \textbf{Avg. Time} & 127.64 & \textbf{34.50} & 86.97 & 48.05 & 245.20 & 86.68 & \underline{39.88} & 261.11 & 86.51 \\
			& \textbf{Avg. Rank} & 5.29 & 5.49 & \underline{3.87} & 5.90 & 4.15 & 5.55 & 6.33 & 5.74 & \textbf{2.73} \\
			\bottomrule
		\end{tabular}
	}
\end{table*}

\subsubsection{Comparison on CMOP Test Suites}

Table~\ref{tab:summary} summarizes the IGD$^+$ results on 87 instances from seven benchmark suites and 28 real-world problems. The Nemenyi post-hoc test gives a critical difference (CD) of approximately 1.12 at the 0.05 significance level. DCF2D achieves the best overall average rank of 2.73, outperforming the second-ranked IMTCMO (3.87) by 1.14. Since this difference exceeds the CD, the overall advantage of DCF2D is statistically significant.

On DASCMOP, DCF2D achieves the best overall performance. The constraints in this suite play diverse roles: some render the UPF infeasible and shift the CPF away from it; some fragment the CPF into disconnected segments; and others merely obstruct the evolutionary path, reshape only part of the CPF, make the CPF dependent on the corresponding RCPF, or have no direct effect on the CPF. Although most algorithms can locate the approximate region of the CPF, they often fail to recover all of its disconnected segments. This limitation is particularly evident when the CPF is closely related to the RCPF, where DCF2D's bidirectional search provides a clear advantage.

On DOC, DCF2D also ranks first. Although DOC problems contain many constraints, only a subset of them is closely related to the final CPF. DCF2D dynamically identifies these critical constraints and allocates auxiliary populations to explore their boundaries, avoiding the unnecessary evaluations incurred by methods that treat all constraints equally.

DCF2D exhibits the best and most stable performance on CDTLZ. These problems typically have a CPF separated from the UPF, and the search toward the CPF encounters numerous local optima and disconnected infeasible barriers in the objective space. In Stage 1, DCF2D searches without being restricted by these disconnected infeasible regions, although it may occasionally converge to local optima. Stage 2 then introduces reverse-direction search, providing additional pathways to cross infeasible regions induced by different constraints and guide the population toward the global optimum.

On LIRCMOP, DCF2D remains highly competitive and clearly outperforms most unidirectional and decoupling-based methods. Because the CPF is often adjacent to large infeasible regions, conventional evolutionary-direction search can be easily obstructed. DCF2D instead approximates the RCPF from the infeasible side, turning these barriers into useful guidance toward the CPF.

On MW, DCF2D outperforms most constraint-decoupling methods but is inferior to APSEA. A possible reason is that the geometric structures and local optima of MW problems favor the genetic operators and adaptive population-sizing strategy used by APSEA over the differential evolution operators adopted by DCF2D. Nevertheless, DCF2D's advantage over other decoupling methods supports the effectiveness of its bidirectional search strategy.

On ZXH\_CF, DCF2D remains competitive with most decoupling-based methods but performs worse than ACREA. The correlated position and distance variables in ZXH\_CF produce narrow feasible corridors, for which ACREA's dual-archive and adaptive constraint-relaxation mechanisms are particularly effective. Despite this limitation, DCF2D benefits from its bidirectional search and Stage 3 refinement in comparison with other decoupling methods.

On SDC, DCF2D ranks behind IMTCMO but still outperforms most decoupling-based competitors. Since many SDC constraints are defined in the decision space, their mappings to the objective space are highly complex, reducing the effectiveness of objective-space decoupling. Even under these conditions, the dynamic bidirectional mechanism of DCF2D remains more robust than static or fuzzy decoupling strategies.

\begin{figure}[htbp]
    \centering
    \includegraphics[width=8.5cm]{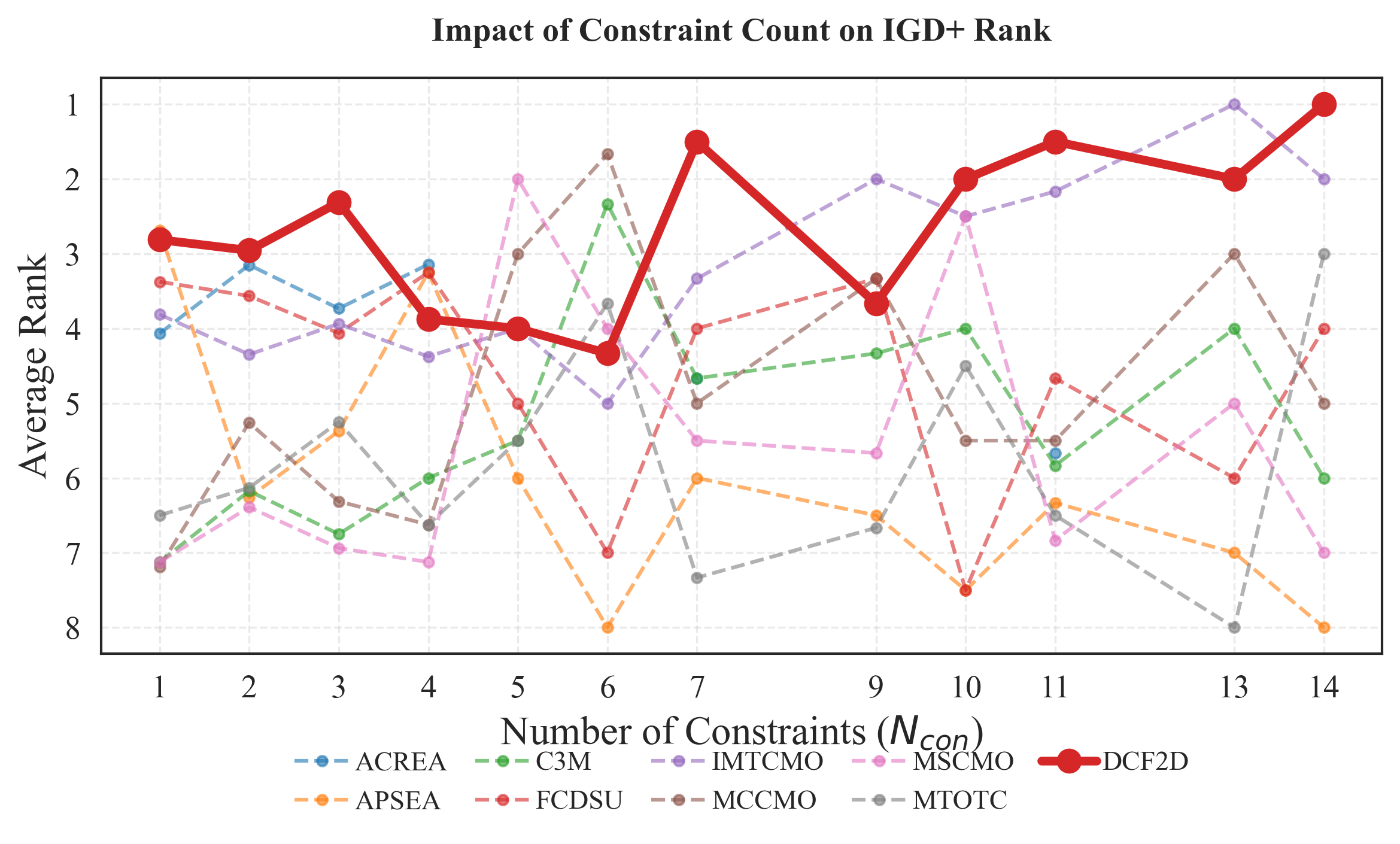}
    \caption{Average IGD$^+$ ranks of DCF2D and the compared algorithms with respect to the number of constraints.}
    \label{numberofcon}
\end{figure}

We further examine the effect of the number of constraints, denoted by $N_{con}$, on the average IGD$^+$ rank in Fig.~\ref{numberofcon}. DCF2D maintains stable performance for problems with low to moderate numbers of constraints and becomes increasingly competitive as $N_{con}$ grows. In particular, it achieves the best or near-best ranks for most settings with $N_{con}\geq 7$ and ranks first when $N_{con}=14$.
This trend indicates that bidirectional constraint decoupling is particularly effective for highly constrained problems. As the number of constraints increases, the CPF is more likely to be bounded or obstructed by interacting infeasible regions. In such cases, reverse-direction search complements conventional evolutionary-direction search by exploiting informative boundaries on the infeasible side.

\subsubsection{Comparison on Real-World CMOPs}

Real-world CMOPs (RWMOPs) commonly involve highly nonlinear constraints that induce narrow or disconnected feasible regions. As summarized in Table~\ref{tab:summary}, decoupling-based methods generally perform better than aggregation-based methods on these problems. One possible reason is the substantial scale disparity among real-world constraints: directly aggregating constraint violations may cause large-magnitude constraints to dominate smaller but equally important ones, whereas constraint decoupling evaluates each constraint separately and is therefore less sensitive to scale differences.

DCF2D achieves the best overall performance on the RWMOP suite, with an average HV rank of 3.46. Its advantage mainly arises from the diversity maintained by the bidirectional search in Stage 2. By simultaneously approximating the SCPF and RCPF, DCF2D preserves boundary solutions from complementary directions and reduces the risk of becoming trapped in isolated feasible regions. Unidirectional decoupling methods may overlook the ICPF, whereas the reverse-direction search of DCF2D exploits the RCPF as an intermediate geometric guide toward the ICPF.

\subsection{Computational Complexity Analysis}

The computational complexity of DCF2D is dominated by environmental selection. In \textbf{Stage 1}, both $MP$ and $AP_0$ maintain $N$ solutions and are updated using candidate sets of size $3N/2$.
Therefore, their total per-generation complexity is $O(MN^2)$, where $M$ and $N$ denote the number of objectives and the population size, respectively.

In \textbf{Stage 2}, updating $MP$ and $LI$ through nondominated sorting requires $O(MN^2)$ time. Let $n_{\mathrm{act}}$ be the number of active constraint-specific auxiliary populations. Because they share $N/2$ offspring, each $AP_i$ selects from approximately $N/4+N/(2n_{\mathrm{act}})$ candidates. Assuming $O(S\log S)$ complexity for the simplified $k$-NN-based selection on $S$ candidates, updating
all active auxiliary populations requires $O(n_{\mathrm{act}}N\log N)$ time. Thus, the per-generation complexity of Stage 2 is
\begin{equation}
    O\left(MN^2+n_{\mathrm{act}}N\log N\right).
\end{equation}
Since $n_{\mathrm{act}}\leq n_{con}$, its worst-case complexity is $O(MN^2+n_{con}N\log N)$. In practice, event-driven activation and
deactivation generally keep $n_{\mathrm{act}}$ well below $n_{con}$.

Although the constant factor does not alter the asymptotic complexity, the reduced auxiliary-population size substantially lowers the actual computational cost. Thus, the marginal cost of activating an $AP_i$ is considerably lower in practice than that of maintaining an additional full-sized population with nondominated sorting. By contrast, multi-population methods such as MTOTC and MCCMO generally maintain a full-sized population for each decoupled task, leading to a complexity that can scale as $O(n_{con}MN^2)$.

\begin{figure}[htbp]
    \centering
    \includegraphics[width=\linewidth]{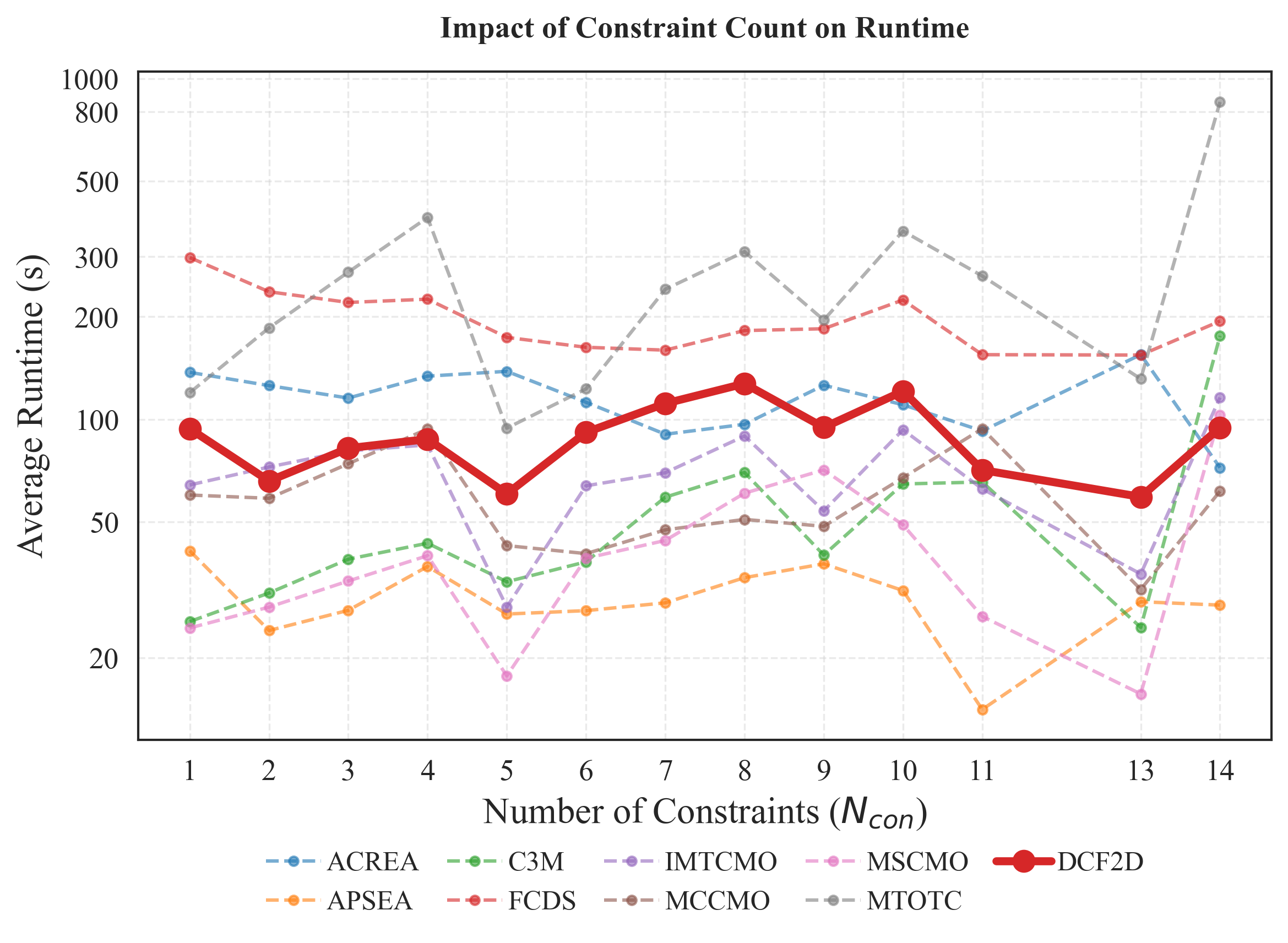}
    \caption{Average runtime of DCF2D and the compared algorithms with respect to the number of constraints ($n_{con}$).}
    \label{fig:runtime_trend}
\end{figure}

The empirical runtime results in Fig.~\ref{fig:runtime_trend} are consistent with the above analysis. Although the runtime of DCF2D generally increases with $n_{con}$, it does not grow proportionally to the total number of constraints; for example, its runtime at $n_{con}=11$ is lower than that at $n_{con}=8$ and $n_{con}=10$. This non-monotonic variation can be attributed to the event-driven decoupling mechanism of DCF2D. Specifically, auxiliary populations are activated only for constraints detected as relevant to the CPF through their SCPFs or RCPFs and are terminated once they no longer provide useful search information. Consequently, redundant constraints, constraints unrelated to the CPF, and auxiliary populations that have completed their search tasks incur little or only temporary population-maintenance cost. By allocating computational resources only to active and informative constraint-wise searches, DCF2D maintains a runtime comparable to, and often lower than, that of other decoupling-based algorithms while remaining substantially more efficient than static decoupling methods such as MTOTC and FCDS.

\subsection{Ablation Study and Parameter Sensitivity}

We conducted ablation experiments and a sensitivity analysis to evaluate the bidirectional search strategy, the multi-stage framework, and the stage-switching parameter $\beta$. The results are summarized in Table~\ref{tab:ablation_beta}.

Two unidirectional variants were considered: \textbf{DCF2D\_EVO}, which performs constraint decoupling only in the evolutionary direction to approximate the SCPF, and \textbf{DCF2D\_REV}, which searches only in the reverse direction to approximate the RCPF. DCF2D achieves the best average rank of 2.95, outperforming DCF2D\_EVO (4.60) and DCF2D\_REV (4.36). This confirms that the SCPF and RCPF provide complementary guidance and that neither direction alone can adequately address the diverse constraint geometries of CMOPs.

We further varied $\beta$, which specifies the fraction of the evaluation budget used before auxiliary populations are deactivated and Stage 3 begins. The best average rank is obtained at $\beta=0.9$, whereas $\beta=1.0$, which removes Stage 3, yields the worst rank of 5.33. Without Stage 3, auxiliary populations may continuously inject mutually nondominated but locally clustered solutions into the main population. This effect is particularly pronounced for problems with more objectives, where a large proportion of solutions may lie on the first nondominated front, weakening density-based truncation and impairing global distribution. Stage 3 avoids this interference by terminating external injection and allowing the main population to refine convergence and diversity independently.

Conversely, a small $\beta$ may terminate auxiliary populations before the decoupled constraint boundaries are sufficiently explored. Although the results are not strictly monotonic, larger values generally perform better: the average ranks for $\beta=0.3$, $0.5$, $0.7$, and $0.9$ are 3.92, 3.20, 3.64, and 2.95, respectively. These results suggest that Stage 2 primarily supports exploration, whereas Stage 3 provides final exploitation. Accordingly, $\beta=0.9$ offers the best balance by allocating most evaluations to bidirectional decoupling while reserving 10\% for final refinement.

A fixed threshold is adopted because a reliable adaptive transition criterion is difficult to construct from local population states. In particular, temporary stagnation may indicate either convergence to the CPF or entrapment in a local region. The fixed setting therefore prevents Stage 2 from being terminated prematurely and guarantees a predefined exploration budget.

\begin{table}[htbp]
    \centering
    \caption{Average ranks of the search-direction variants and different settings of $\beta$ (lower is better).}
    \label{tab:ablation_beta}
    \resizebox{8.5cm}{!}{
        \begin{tabular}{llc}
            \toprule
            \textbf{Category} & \textbf{Variant / Setting} & \textbf{Avg. Rank} \\
            \midrule
            \multirow{3}{*}{\textbf{Search Direction}}
            & DCF2D\_EVO (Evolutionary Only) & 4.60 \\
            & DCF2D\_REV (Reverse Only) & 4.36 \\
            & \textbf{DCF2D (Bidirectional)} & \textbf{2.95} \\
            \midrule
            \multirow{5}{*}{\textbf{Parameter $\beta$}}
            & $\beta=0.3$ & 3.92 \\
            & $\beta=0.5$ & 3.20 \\
            & $\beta=0.7$ & 3.64 \\
            & \textbf{$\beta=0.9$ (Default)} & \textbf{2.95} \\
            & $\beta=1.0$ (No Stage 3) & 5.33 \\
            \bottomrule
        \end{tabular}
    }
\end{table}

\section{Conclusion}

This study investigates the geometric roles of individual constraints in evolutionary constrained multi-objective optimization. By defining the SCPF, ICPF, and RCPF, we provide a white-box interpretation of how individual constraints and their interactions shape the CPF. The analysis shows that feasible-side information alone may be insufficient in highly constrained landscapes; RCPFs can also provide valuable intermediate guidance toward difficult-to-reach optimal regions. Based on this insight, DCF2D employs bidirectional constraint decoupling to exploit complementary guidance from SCPFs and RCPFs. The experimental results demonstrate its effectiveness, particularly on problems with disconnected feasible regions or extensive infeasible barriers.

DCF2D nevertheless has two main limitations. First, it decouples constraints individually and may therefore overlook interactions among composite or strongly coupled constraints, leading to redundant exploration. Second, the Pareto-dominance-based screening of the infeasible archive ($LI$) may lose selection pressure as the number of objectives increases.
Future work will investigate automatic constraint grouping to coordinate strongly coupled constraints, as well as indicator- or reference-vector-based archive management for constrained many-objective optimization. Another promising direction is dynamic constrained multi-objective optimization, where adaptive activation and direction switching could be used to track time-varying relationships among constraint-induced boundaries.

\bibliography{reference}
\end{document}